\documentclass{article}

\PassOptionsToPackage{numbers, compress}{natbib}

\usepackage[preprint]{neurips_2026}

\usepackage{enumitem}

\usepackage{hyperref}       
\usepackage{url}            
\usepackage{booktabs}       
\usepackage{amsfonts}       
\usepackage{nicefrac}       
\usepackage{microtype}      
\usepackage{xcolor}         
\usepackage{amsmath}
\usepackage{booktabs}
\usepackage{graphicx}
\usepackage{multirow}
\usepackage{float}

\usepackage{iftex}

\ifPDFTeX
  \usepackage[utf8]{inputenc}
  \usepackage[T1]{fontenc}
\else
  \usepackage{fontspec}
  \usepackage[UTF8]{ctex}
\fi

\title{Geo-Align: Video Generation Alignment via Metric Geometry Reward}

\author{%
  Zizun Li$^{1,2}$\quad
  Haoyu Guo$^{2}$\thanks{Corresponding author.}\quad
  Runzhe Teng$^{1,2}$\quad
  Chunhua Shen$^{2,3}$ \quad
  Tong He$^2$\quad
  \vspace{0.15cm} \\
  \textmd{$^1$USTC \quad $^2$Shanghai AI Lab \quad $^3$ZJU}\\
  \vspace{-0.15cm} \\
  \textmd{\url{https://lizizun.github.io/geo-align-page/}}
}

\begin{document}

\maketitle

\vspace{-0.6cm}
\begin{figure}[H]
  \centering
  \includegraphics[width=0.99\linewidth]{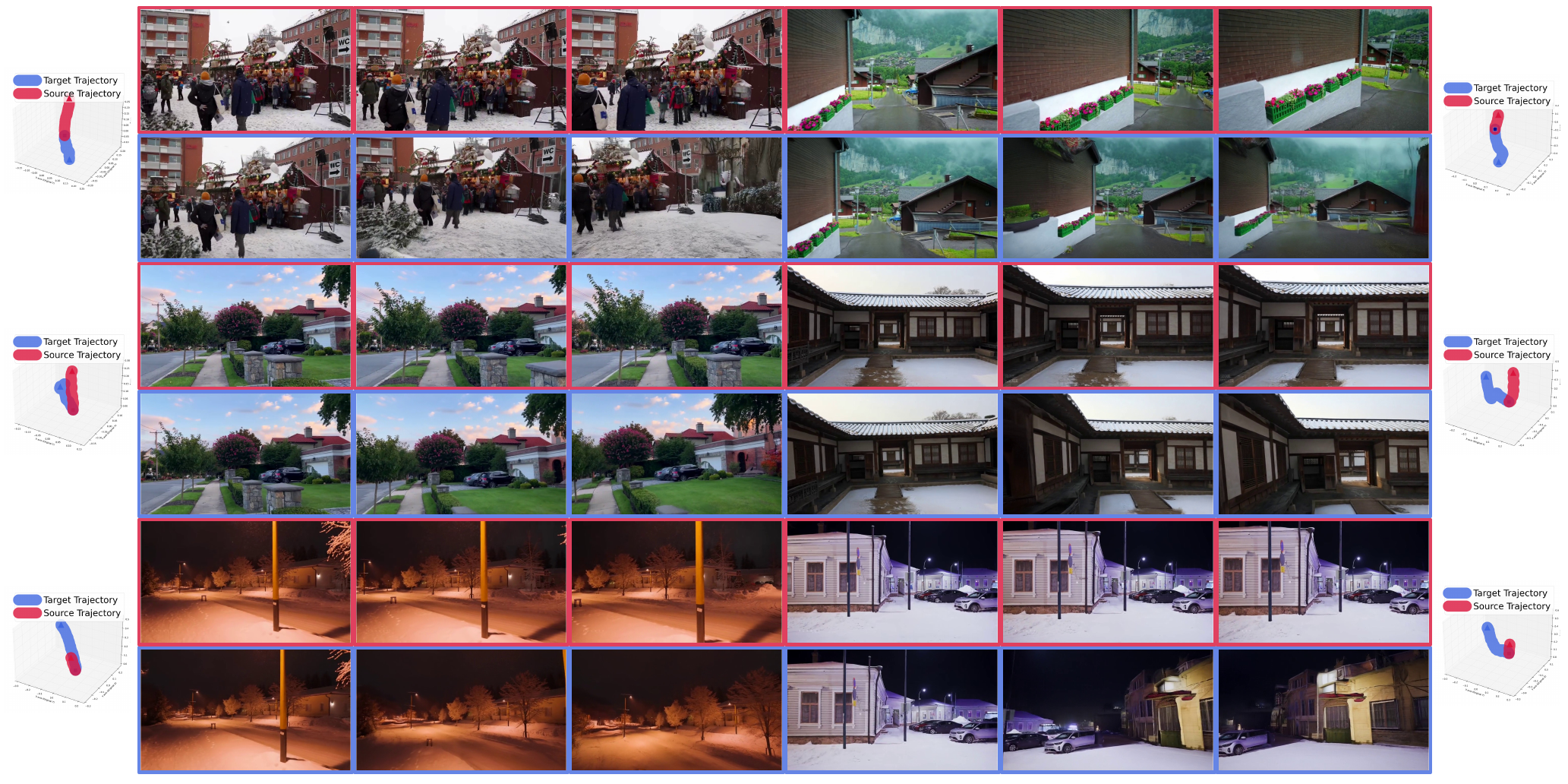}
  \caption{Given a conditioning video, Geo-Align synthesizes a novel view video according to the target camera trajectory.}
  \label{fig:teaser}
\end{figure}

\begin{abstract}
  Camera-controlled video generation has achieved remarkable progress in recent years. However, existing video-to-video re-rendering methods primarily rely on Supervised Fine-Tuning using synthetic datasets. At present, there is an extreme scarcity of synchronized, multi-view real-world video data. Consequently, the prevailing paradigm often exhibits limited generalization when processing out-of-distribution real-world videos, with models struggling to accurately adhere to physical scales and camera trajectories. To bridge this gap, we propose Geo-Align, the first Reinforcement Learning framework specifically designed for camera-controlled video re-rendering. Built upon a pretrained model, we optimize the model through a scale-aware perceptual reward mechanism. Specifically, we introduce a metric 3D estimator to extract precise camera trajectories from generated videos, explicitly penalizing deviations in rotation and translation. Furthermore, we meticulously designed a data pipeline strategy based on real-world conditioning videos and target camera trajectories derived from synthetic data, eliminating the reliance on paired data. Extensive experiments demonstrate that Geo-Align consistently outperforms existing supervised learning baselines in both precise camera controllability and visual fidelity, indicating the effectiveness of our method.
\end{abstract}

\section{Introduction}

Camera controllability plays a vital role in video generation, particularly in fields such as film production and game engine rendering.
In this paper, we focus on the video retake task. Formulated as a video-to-video generation problem, this task requires a model to synthesize a novel-view video along a target camera trajectory, given a conditioning video and the target trajectory as inputs. Recent methods such as ReCamMaster \citep{bai2025recammaster} and ReDirector \citep{park2025redirector}, have successfully re-rendered dynamic scenes from input videos along new camera trajectories by training on synthetic datasets generated via engines like Unreal Engine.
While TrajectoryCrafter \citep{yu2025trajectorycrafter} and CogNVS \citep{chen2025reconstruct} are methods based on reconstruction, warping, and subsequent completion. However, the current supervised learning paradigm for generating videos with novel camera trajectories faces two core bottlenecks:

\textbf{Data Scarcity}: Unlike camera-controlled video generation conditioned on a single initial frame, video retake requires multi-view video data for supervised training. Given the scarcity of such real-world data, implicit condition methods \citep{bai2025recammaster,park2025redirector} predominantly rely on synthetic datasets, while warping-based methods \citep{yu2025trajectorycrafter,van2024generative,chen2025reconstruct} rely on point cloud renderings to synthesize target videos, constructing such data is highly non-trivial.
While fine-tuning on synthetic data yields impressive results, these models often exhibit significant domain shift when performing inference on real-world scenes.

\textbf{Metric Ambiguity}: Camera pose annotations for existing real-world videos are often scale-less. Even the MultiCam-Video data constructed by ReCamMaster \citep{bai2025recammaster} only provides metric information for synthetic data. Standard SFT loss functions focus on pixel-level or feature-level reconstruction rather than explicitly optimizing for physically meaningful, metric-level camera alignment, frequently leading to scale drift in generated trajectories.

To address these challenges, we propose Geo-Align, a framework that introduces Reinforcement Learning (RL) to directly optimize the physical alignment and visual quality of camera movements. Unlike previous SFT paradigms \citep{bai2025recammaster,park2025redirector} that rely on time-synchronized ground-truth videos from multiple camera angles, reinforcement learning methods do not require video data corresponding to the target camera trajectory. Since real-world conditioning videos are easily obtainable, we can post-train the model via RL as long as we have the target camera trajectory.
We adopt a fusion strategy combining real and synthetic data. During RL training, the conditioning videos are real-world captures. For the target camera trajectories, we sample from OmniWorld \citep{zhou2025omniworld} gaming data, which provides a rich variety of natural camera movements. Since gaming trajectories are typically non-metric, we perform rescaling using Truncated Gaussian Sampling. Specifically, we sample the maximum values for rotation and translation between adjacent frames within defined thresholds and rescale the camera trajectories to reasonable scales accordingly.

We utilize a Verifiable Geometry Reward to train our model, which compares the camera trajectories estimated from the generated video (via MapAnything \citep{keetha2025mapanything}) against the target trajectories. A metric evaluator is introduced to mitigate metric-related reward hacking during the reinforcement learning process. This effectively penalizes degenerate solutions—such as the model producing a shape-preserving but slow-moving trajectory in response to a rapid target trajectory. To prevent visual degradation during geometric optimization and preserve the model’s priors, we also incorporate aesthetic rewards, utilizing VideoAlign \citep{liu2025improving} and HPSv3 \citep{ma2025hpsv3} as the reward models. We freeze the majority of the model's parameters, training only the self-attention layers.

We evaluate our model on the DAVIS \citep{pont20172017} datasets across the ten target camera trajectory categories defined by ReCamMaster \citep{bai2025recammaster}. Results demonstrate that our RL-trained model not only improves accuracy in following target trajectories on real-world data but also outperforms the original model across various aesthetic evaluation metrics. Our core contributions are as follows:

\begin{itemize}[leftmargin=1.5em, labelsep=0.5em, itemindent=0pt, labelindent=0.5em]

\item \textbf{Reinforcement Learning for Video Retake}: We utilize metric geometry model to extract rotation and translation errors. This enables our model to better align with geometric constraints and achieve more accurate metric scaling in real-world conditioning videos. Furthermore, we incorporate aesthetic rewards to enhance the overall quality of the generated videos.

\item \textbf{Fusion Data Strategy}: We leverage MapAnything \citep{keetha2025mapanything} to extract camera poses from Citywalk \citep{li2025sekai} dataset as real-world conditioning priors. By combining this with Truncated Gaussian Sampling to rescale target trajectories from gaming data, we enhance training diversity and bridge the scale gap between source videos and target trajectories. Furthermore, it circumvents the necessity of paired multi-view video data.

\item \textbf{State-of-the-Art (SOTA) Performance}: Our RL-trained model achieves SOTA performance on the DAVIS \citep{pont20172017} dataset across ReCamMaster's \citep{bai2025recammaster} 10 trajectory types, consistently improving both camera trajectory fidelity and overall visual aesthetics. Qualitative comparisons further demonstrate a noticeable improvement in the quality of the generated videos.
\end{itemize}

\section{Related Work}

\subsection{Camera-Controlled Video Retake}

Camera-controlled video retake \citep{bahmani2025ac3d, he2024cameractrl, wang2024motionctrl, go2025splatflow} aims to synthesize novel views from existing footage by redirecting camera trajectories through generative models. Early approaches predominantly rely on explicit geometric transformations, utilizing external depth estimators \citep{hu2025depthcrafter, chen2025video} and point trackers \citep{karaev2024cotracker, xiao2024spatialtracker} to warp input frames before refining them with video diffusion models \citep{wan2025wan, kong2024hunyuanvideo, yang2024cogvideox}, as seen in methods like TrajectoryCrafter \citep{yu2025trajectorycrafter} and CogNVS \citep{chen2025reconstruct}. However, these explicit methods frequently suffer from warping artifacts that propagate directly into the synthesized output, particularly under dynamic camera motions or complex scene structures. To bypass explicit warping, implicit methods \citep{chen2025reconstruct, jeong2025reangle, lu2025see4d, zhang2025recapture} such as Generative Camera Dolly (GCD) \citep{van2024generative} and ReCamMaster \citep{bai2025recammaster} condition models directly on camera extrinsic parameters, internalizing multi-view geometry through synthetic datasets. While recent advancements like ReDirector \citep{park2025redirector} extend this implicit paradigm to handle variable-length inputs and dynamic motions via Rotary Camera Encoding (RoCE). ll these frameworks fundamentally rely on supervised fine-tuning, where the primary bottleneck is the severe scarcity of time-synchronized multi-view video data. Since constructing such datasets from real-world footage is exceedingly difficult, existing SFT methods \citep{bai2025recammaster,park2025redirector} are forced to rely heavily on synthetic data.

\subsection{Feed-Forward 3D Reconstruction}

Recent feed-forward models directly predict scene geometry without traditional SfM optimization \citep{triggs1999bundle, snavely2006photo, wu2013towards, schoenberger2016sfm}. DUSt3R \citep{wang2024dust3r} pioneered this by regressing dense point maps from unconstrained images. To handle continuous visual streams, methods \citep{lan2025stream3rscalablesequential3d,wang20243dreconstructionspatialmemory,zhuo2026streaming4dvisualgeometry,liu2025slam3rrealtimedensescene} such as CUT3R \cite{wang2025continuous} and WinT3R \citep{li2025wint3r} introduced stateful memory and sliding-window mechanisms for efficient online perception. Concurrently, models \citep{zhang2026flarefeedforwardgeometryappearance,yang2025fast3r3dreconstruction1000,deng2026vggtlongchunkitloop,leroy2024groundingimagematching3d,zhang2025monst3rsimpleapproachestimating} like VGGT \citep{wang2025vggt}, $\pi^3$ \citep{wang2025pi}, and Depth Anything 3 \citep{lin2025depth} have scaled into unified foundational architectures capable of jointly inferring multi-view geometry, cameras, and depth. Despite these advances, achieving accurate metric-scale reconstruction remains challenging. To address this, MapAnything \citep{keetha2025mapanything} introduces a universal framework specifically for metric 3D reconstruction. By employing a factored representation that decouples camera poses and depth into scale-invariant components and explicit global scales, MapAnything \citep{keetha2025mapanything} robustly maps local geometry into a unified metric space without test-time optimization.

\subsection{Group Relative Policy Optimization in Generative Models}

Group Relative Policy Optimization (GRPO) \citep{guo2025deepseek} has recently emerged as a powerful online reinforcement learning framework for aligning generative models \citep{fei2025srposelfreferentialpolicyoptimization,geng2025xomnireinforcementlearningmakes,xue2025dancegrpounleashinggrpovisual}. In flow-matching \citep{lipman2022flow} domains, Flow-GRPO \citep{liu2025flow} enables online RL via ODE-to-SDE conversion, while MixGRPO \citep{li2025mixgrpo} further improves optimization efficiency by introducing a mixed ODE-SDE sliding window sampling mechanism. This paradigm has similarly advanced video generation: GrndCtrl \citep{he2025grndctrl} utilizes GRPO for physically grounded world modeling, and recent frameworks \citep{wang2025taming, ge2026campilotimprovingcameracontrol, wang2026worldr1reinforcing3dconstraints} adopt verifiable geometry rewards to optimize precise camera-controlled video generation. Another line of work enhance synthesis quality by incorporating explicit \citep{wu2025icworldincontextgenerationshared, kupyn2025epipolargeometryimprovesvideo, yin2026vigorvideogeometryorientedreward} or implicit \citep{an2026vggrpoworldconsistentvideogeneration, yan2025rlgfreinforcementlearninggeometric, du2026videogpadistillinggeometrypriors, wu2025geometryforcingmarryingvideo} geometric constraints as reward signals to enforce multi-view consistency.
Furthermore, LongCat-Video \citep{team2025longcat} demonstrates robust multi-reward RLHF in foundational video models by introducing crucial stabilization techniques—specifically, employing max group standard deviation to bound reward variances within groups, and utilizing policy and KL loss reweighting to dynamically balance optimization and prevent reward hacking. Building upon these advancements, our method synergistically integrates the efficient mixed sampling framework of MixGRPO \citep{li2025mixgrpo} with the max group standard deviation and policy/KL loss reweighting strategies from LongCat-Video \citep{team2025longcat}, achieving highly stable and computationally efficient policy optimization.

\section{Methodology}

\subsection{Overview}

Given an input conditioning video and a user-specified, unseen camera trajectory, we aim to re-render and generate a novel view video sequence.
Formally, let $\mathbf{x}_{1:N}$ denote the conditioning video of length $N$, and $c$ be the corresponding text prompt. 
To guide the generation process along a designated path, the model is additionally conditioned on a target camera trajectory $\mathbf{P}^{tgt}_{1:N}$, including target camera intrinsic parameters $\mathbf{K}_{1:N}^{tgt}$ and extrinsic parameters $\mathbf{E}_{1:N}^{tgt}$.

Our framework is built upon a pretrained video world model, denoted as $\mathcal{W}_\theta$. During the iterative generation process (e.g., diffusion or flow matching), the model predicts the denoised representation (or velocity vector) given a noisy latent $\mathbf{z}_t$ at timestep $t$. The conditional generation process can be formulated as:
\begin{equation}
\hat{\mathbf{v}}_\theta = \mathcal{W}_\theta(\mathbf{z}_t, t, \mathcal{C}),
\end{equation}
where $\mathcal{C} = \{ \mathbf{x}_{1:N}, c, \mathbf{P}^{tgt}_{1:N} \}$ encapsulates all the multimodal conditioning signals.
Although fine-tuning video foundation models multi-view videos offers a viable solution to this task, the inherent scarcity of such data remains a significant bottleneck.
Relying solely on supervised fine-tuning often leads to geometric inconsistencies and suboptimal camera control.
Therefore, the adoption of RL frees us from multi-view data dependencies, unlocking the potential to train on vastly larger and more diverse data.
Our goal is to optimize the model parameters $\theta$ to maximize a composite reward function $\mathcal{R}$, which comprehensively evaluates the alignment between the generated video $\mathbf{y}_{1:N}$ and the target trajectory $\mathbf{P}^{tgt}_{1:N}$, as well as the overall video quality. The RL objective is defined as:
\begin{equation}
\max_{\theta} \mathbb{E}_{\mathbf{y}_{1:N} \sim \mathcal{W}_\theta(\cdot | \mathcal{C})} \left[ \mathcal{R}(\mathbf{y}_{1:N}, \mathbf{P}^{tgt}_{1:N}) \right].
\end{equation}
By directly optimizing this reward, the model is guided to strictly adhere to the prescribed target trajectory while maintaining superior spatiotemporal fidelity.

\begin{figure}
  \centering
  \includegraphics[width=0.99\linewidth]{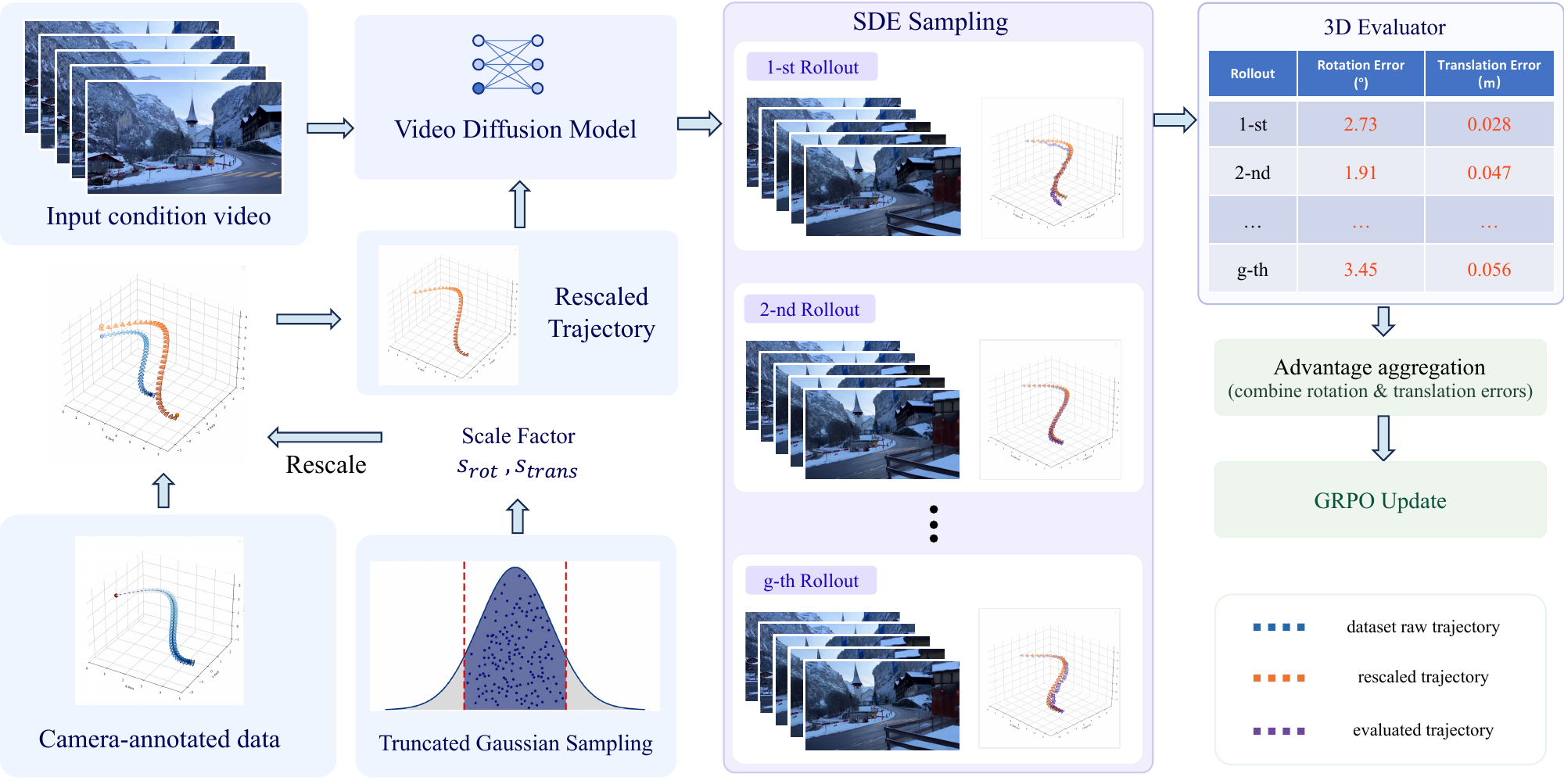}
  \caption{\textbf{Geo-Align pipeline.} Given a conditioning video, we sample a camera trajectory from other camera-annotated data and scale it to a plausible range, with the scaling factor drawn from a truncated Gaussian distribution. After the model generates a set of rollout videos, a metric 3D evaluator assesses the camera trajectory of each sample to compute geometry rewards. Finally, the model is optimized via Group Relative Policy Optimization \citep{guo2025deepseek}.}
  \label{fig:pipeline}
\end{figure}

\subsection{Multi-Dimensional Reward Design}

\textbf{Verifiable Geometry Reward}. To enforce rigorous spatial alignment between the generated video $\mathbf{y}_{1:N}$ and the designated target trajectory, we introduce a verifiable Geometry Reward. We construct our 3D evaluator upon MapAnything \citep{keetha2025mapanything}, a metric feed-forward 3D reconstruction model. By feeding the generated video into the 3D evaluator, we extract the predicted camera trajectory, comprising translations $\hat{\mathbf{t}}_{1:N}$ and rotations $\hat{\mathbf{R}}_{1:N}$. The geometric discrepancy is then quantified against the input target trajectory $(\mathbf{t}_{1:N}, \mathbf{R}_{1:N})$ across two dimensions. Specifically, we compute the weighted Euclidean deviation for translation and the angular deviation for rotation:
\begin{equation}
D_{trans} = \sum_{i=1}^N w_i \|\mathbf{t}_i - \hat{\mathbf{t}}_i\|_2,
\end{equation}
\begin{equation}
D_{rot} = \sum_{i=1}^N w_i \arccos\left(\frac{\text{Tr}(\mathbf{R}_i^\top \hat{\mathbf{R}}_i) - 1}{2}\right),
\end{equation}
where $w_i$ represents the temporal weight for the $i$-th frame. A key empirical observation motivates this weighting scheme: pretrained video generative models typically exhibit strong adherence to the conditioning trajectory in the initial frames, but suffer from severe error accumulation and spatial drift in the latter frames. Because the latter frames more accurately reflect the model's true predictive capability and are the primary bottleneck in trajectory control, we design $w_i$ as a monotonically increasing function of time $i$ (e.g., $w_1 < w_2 < \dots < w_N$). This temporally progressive weighting mechanism explicitly penalizes long-term drift and forces the RL process to prioritize the optimization of challenging latter frames.

\textbf{Perceptual and Aesthetic Reward}. Optimizing solely for geometric alignment can inadvertently lead to reward hacking, resulting in perceptual degradation or unnatural artifacts. To preserve and enhance the visual fidelity of the synthesized video, we incorporate multidimensional aesthetic and quality rewards. First, we leverage the VideoAlign \citep{liu2025improving} evaluator to assess sequence-level dynamics, yielding a visual quality score ($s_{vis}$) and a motion quality score ($s_{mot}$). Furthermore, to guarantee superior single-frame visual aesthetics and high-frequency details, we utilize HPSv3 \citep{ma2025hpsv3} to evaluate the perceptual quality of each individual frame. 

\subsection{Flow Matching Optimization via GRPO}

To efficiently optimize the pretrained flow matching model $\mathcal{W}_\theta$ for trajectory-controlled generation, we employ Group Relative Policy Optimization \citep{guo2025deepseek}. Traditional PPO \citep{schulman2017proximalpolicyoptimizationalgorithms} relies on a memory-intensive value model for baseline estimation. GRPO resolves this memory constraint by removing the value model and leveraging the relative scores within a group of outputs to compute the advantage.
Given the prohibitively long group sampling time of video generation models, we adopt the sliding-window sampling strategy from MixGRPO \citep{li2025mixgrpo}. This mechanism restricts stochastic sampling and gradient updates strictly to an active temporal window, significantly accelerating convergence. Furthermore, since directly summing multi-dimensional rewards is mathematically unstable, we aggregate the feedback in the advantage space. Following the max group standard deviation strategy (as in LongCat Video \citep{team2025longcat}), we robustly normalize each reward dimension $k \in \{rot, trans, vis, mot, hps\}$ within a group of $G$ sampled rollouts to prevent the over-amplification of low-variance noise:
\begin{equation}
\hat{A}_k^{(j)} = \frac{r_k^{(j)} - \mu_k}{\max(\sigma_k, \epsilon)},
\end{equation}
where $\mu_k$ and $\sigma_k$ are the group mean and standard deviation. The total advantage $A_{total}^{(j)}$ is then formulated as:
\begin{equation}
A_{total}^{(j)} = \sum_{i \in k}\lambda_{i} \hat{A}_{i}^{(j)}.
\end{equation}
Standard GRPO incorporates a KL-divergence penalty to anchor the policy to the pretrained model. However, to maximize the model's exploratory capability on entirely novel, out-of-distribution target camera trajectories, we remove this KL penalty. Incorporating a timestep-aware policy loss weight $w_t$ to balance gradients across diffusion stages as in LongCat Video \citep{team2025longcat}, our final objective function is:
\begin{equation}
\mathcal{J}(\theta) = \mathbb{E}_{t, \mathbf{z}_t, \mathcal{C}} \left[ \frac{1}{G} \sum_{j=1}^G w_t \min \left( \rho_t^{(j)} A_{total}^{(j)}, \text{clip}\left(\rho_t^{(j)}, 1-\epsilon_c, 1+\epsilon_c\right) A_{total}^{(j)} \right) \right].
\end{equation}
where $\rho_t^{(j)}$ denotes the policy probability ratio, and $\epsilon_c$ is the clipping hyperparameter.

\subsection{Metric-Aware Data Sampling Pipeline}

Benefiting from the RL framework, our approach eliminates the reliance on paired ground-truth videos, unlocking the ability to train on large-scale, in-the-wild data.

Specifically, for the conditioning inputs, we utilize in-the-wild CityWalk \citep{li2025sekai} videos, which encompass a diverse array of static and dynamic scenes across both indoor and outdoor environments. The source camera trajectories for these uncalibrated conditioning videos are estimated using MapAnything \citep{keetha2025mapanything}. Conversely, to inject a rich and complex repertoire of camera motions into the model, we sample the target trajectories from the OmniWorld \citep{zhou2025omniworld} gaming dataset. However, drawing target trajectories directly from gaming data introduces critical optimization bottlenecks: these trajectories lack an absolute physical metric scale and frequently exhibit severe rotation.
To guarantee the physical plausibility and kinematic stability of the target camera poses during RL training, we introduce a rescaling mechanism. First, we calculate the maximum frame-to-frame translation speed $v_{trans}^{max}$ and rotation speed $v_{rot}^{max}$ of the raw target trajectory $\mathbf{P}^{tgt}_{1:N}$:
\begin{equation}
v_{trans}^{max} = \max_{i \in [1, N-1]} \|\mathbf{t}_{i+1} - \mathbf{t}_i\|_2,
\end{equation}
\begin{equation}
v_{rot}^{max} = \max_{i \in [1, N-1]} \|\log(\mathbf{R}_i^\top \mathbf{R}_{i+1})^\vee\|_2,
\end{equation}
where $\mathbf{t}_i$ and $\mathbf{R}_i$ denote the translation vector and rotation matrix at frame $i$, respectively, and $(\cdot)^\vee$ maps the skew-symmetric matrix in the Lie algebra $\mathfrak{so}(3)$ to its corresponding rotation vector. To ensure the trajectory speeds fall within a reasonable physical bound while maintaining data diversity, we sample target maximum speeds, $\tau_{trans}$ and $\tau_{rot}$, from Truncated Gaussian Distributions:
\begin{equation}
\tau_{trans} \sim \mathcal{N}_{trunc}(\mu_t, \sigma_t^2, [a_t, b_t]),
\end{equation}
\begin{equation}
\tau_{rot} \sim \mathcal{N}_{trunc}(\mu_r, \sigma_r^2, [a_r, b_r]),
\end{equation}
where $[a_t, b_t]$ and $[a_r, b_r]$ define the strict physical bounds for translation and rotation speeds, concentrating the sampling probability around natural human-walking or steady-cam speeds. Finally, we compute the rescaling factors for translation and rotation, denoted as $s_{trans}$ and $s_{rot}$ respectively:
\begin{equation}
s_{trans} = \frac{\tau_{trans}}{v_{trans}^{max} + \epsilon}, \quad s_{rot} = \frac{\tau_{rot}}{v_{rot}^{max} + \epsilon},
\end{equation}
where $\epsilon$ is a small constant to prevent division by zero. The target trajectory is then uniformly rescaled to yield the modified physical-aware trajectory $\tilde{\mathbf{P}}^{tgt}_{1:N}$:
\begin{equation}
\tilde{\mathbf{t}}_i = s_{trans} \mathbf{t}_i,
\end{equation}
\begin{equation}
\tilde{\mathbf{R}}_i = \exp\left(s_{rot} \log(\mathbf{R}_i)\right).
\end{equation}
This rescaling protocol effectively eliminates unnatural camera jumps and aligns the synthetic gaming trajectories with real-world metric scales, significantly stabilizing the RL optimization landscape.

\section{Experiments}

\subsection{Implementation Details}

We adopt ReDirector \citep{park2025redirector} which is based on Wan2.1 \citep{wan2025wanopenadvancedlargescale} 1.3B as our foundational pretrained video generation model. Following our proposed metric-aware data sampling pipeline, we continuously draw conditioning videos from the CityWalk \citep{li2025sekai} dataset and physically rescaled target trajectories from the OmniWorld \citep{zhou2025omniworld} dataset. For the verifiable geometric reward, MapAnything \citep{keetha2025mapanything} is employed as the frozen 3D evaluator. To preserve the strong spatiotemporal prior of the pretrained base model while enabling precise spatial control, we employ a parameter-efficient fine-tuning strategy: during the RL optimization, we solely update the weights of the self-attention layers, keeping all other network components strictly frozen. The model is configured to generate video sequences of $N = 81$ frames at a spatial resolution of $480 \times 832$. During inference and RL sampling, the continuous flow matching generation process is discretized into $T = 25$ denoising timesteps. For the GRPO \citep{guo2025deepseek} reinforcement learning framework, we follow the efficient mixed sampling framework of MixGRPO \citep{li2025mixgrpo} and set the group size to $G = 12$ video rollouts per condition to compute the robust group-normalized advantages. The network is optimized for a total of 140 RL iterations using a constant learning rate of $\eta = 1 \times 10^{-4}$. The post-training process is distributed across 64 NVIDIA A800 GPUs, consuming about 130 hours.

\subsection{Baselines}

We compare our method against two categories of state-of-the-art baselines. The first category comprises explicit warping-based methods, specifically TrajectoryCrafter \citep{yu2025trajectorycrafter} and CogNVS \citep{chen2025reconstruct}. As these models are limited to generating fewer than 49 frames during inference. The second category consists of models conditioned on implicit camera extrinsics, including ReCamMaster \citep{bai2025recammaster} and ReDirector \citep{park2025redirector}, which are capable of generating 81 or more frames.

\begin{figure}
  \centering
  \includegraphics[width=0.99\linewidth]{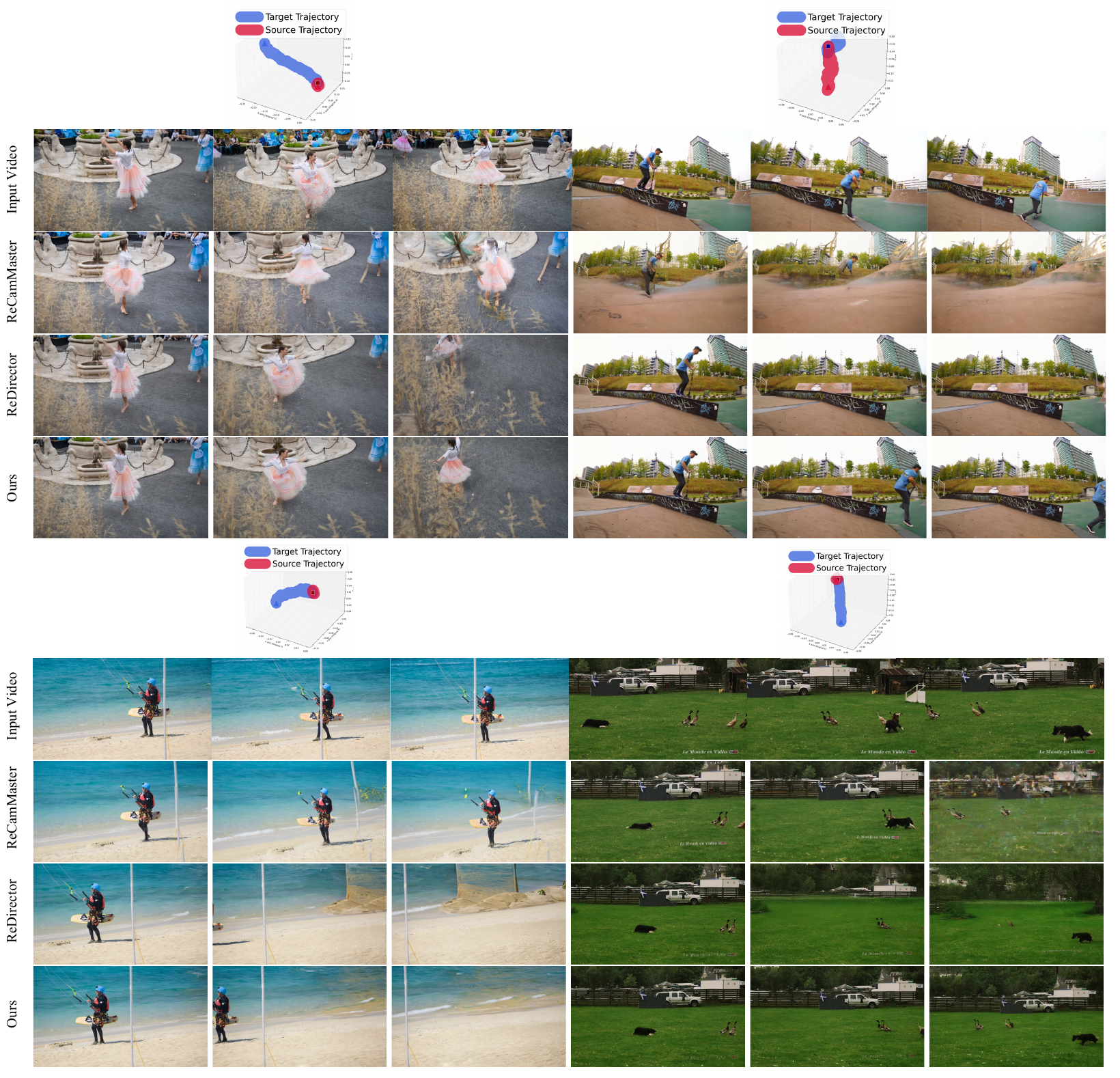}
  \caption{\textbf{Qualitative results on the DAVIS \citep{pont20172017} dataset.} Geo-Align demonstrates superior capabilities in maintaining geometric consistency between the foreground subject and the background, whereas other methods suffer from varying degrees of distortion.}
  \label{fig:Qualitative results}
\end{figure}

\begin{table}
  \caption{\textbf{Quantitative comparison results on different metrics.} Through the proposed reinforcement learning framework, our model yields marked enhancements across all quantitative metrics.}
  \label{tab:performance_comparison}
  \centering
  \renewcommand{\arraystretch}{1.1} 
  \resizebox{\textwidth}{!}{
  \begin{tabular}{llcccccccccc}
    \toprule
    \multirow{3}{*}{Method} & \multirow{3}{*}{Type} & \multicolumn{6}{c}{Visual Quality $\uparrow$} & \multicolumn{2}{c}{Geometric Consistency} & \multicolumn{2}{c}{Camera Accuracy} \\
    \cmidrule(lr){3-8} \cmidrule(lr){9-10} \cmidrule(lr){11-12}
    & & Subject & Background & Aesthetic & Imaging & Temporal & Motion & \multirow{2}{*}{Dyn-MEt3R $\uparrow$} & \multirow{2}{*}{MEt3R $\downarrow$} & \multirow{2}{*}{TransErr $\downarrow$} & \multirow{2}{*}{RotErr $\downarrow$} \\
    & & Consistency & Consistency & Quality & Quality & Flickering & Smoothness & & & & \\
    \midrule
    
    CogNVS \citep{chen2025reconstruct} & explicit warping & 0.9134 & \textbf{0.9342} & 0.4830 & \textbf{0.6398} & \textbf{0.9460} & 0.9768 & 0.8037 & 0.2736 & 0.0367 & \textbf{6.9499} \\
    TrajectoryCrafter \citep{yu2025trajectorycrafter} & explicit warping & \textbf{0.9145} & 0.9331 & \textbf{0.5252} & 0.6394 & 0.9432 & \textbf{0.9800} & \textbf{0.8244} & \textbf{0.2136} & \textbf{0.0293} & 10.4340 \\
    \midrule
    \addlinespace 
    
    Recammaster \citep{bai2025recammaster} & implicit condition & 0.9050 & \textbf{0.9190} & \underline{0.5146} & 0.6611 & \textbf{0.9644} & \textbf{0.9867} & 0.7971 & 0.3485 & 0.0245 & 2.3175 \\
    Redirector \citep{park2025redirector} & implicit condition & \underline{0.9098} & 0.9150 & 0.5141 & \underline{0.6821} & 0.9537 & 0.9856 & \underline{0.8497} & \underline{0.3130} & \underline{0.0149} & \underline{1.4635} \\
    Ours & implicit condition & \textbf{0.9151} & \underline{0.9179} & \textbf{0.5168} & \textbf{0.6842} & \underline{0.9548} & \underline{0.9862} & \textbf{0.8573} & \textbf{0.3077} & \textbf{0.0129} & \textbf{1.3645} \\
    \bottomrule
  \end{tabular}
  }
\end{table}

\begin{figure}[H]
  \centering
  \includegraphics[width=0.99\linewidth]{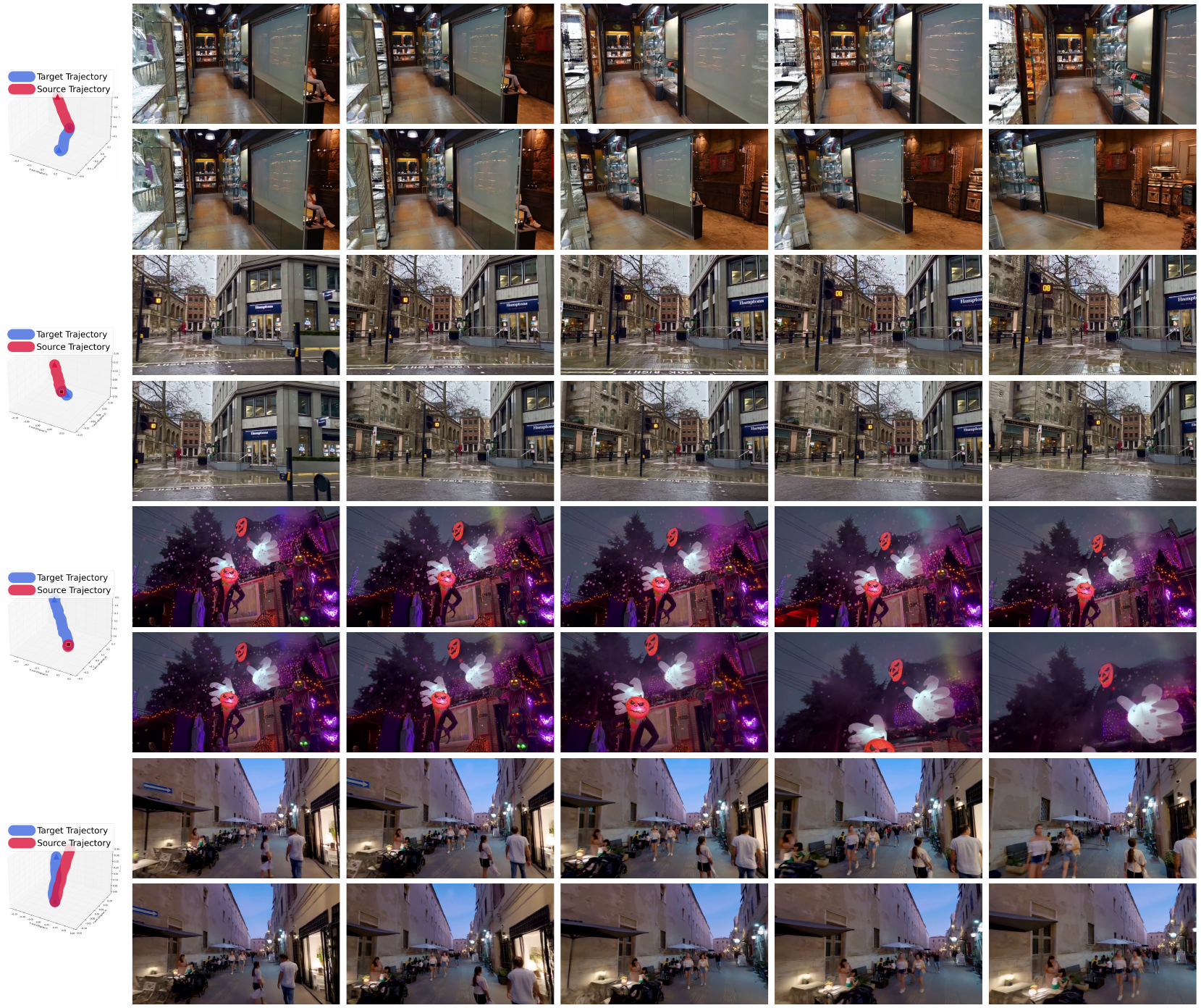}
  \caption{\textbf{More visualization results on CityWalk \citep{li2025sekai} dataset.} For each example, the top row illustrates the input video, whereas the bottom row visualizes our results following the target trajectory.}
  \label{fig:demo2}
\end{figure}

\begin{table}
  \caption{\textbf{Quantitative comparison results across different camera speeds.} Our model consistently outperforms baseline method across varying camera speeds.}
  \label{tab:camera_speed_comparison}
  \centering
  \renewcommand{\arraystretch}{1.1} 
  \resizebox{\textwidth}{!}{
  \begin{tabular}{lccccccccccc}
    \toprule
    \multirow{3}{*}{Method} & \multirow{3}{*}{Camera Speed} & \multicolumn{6}{c}{Visual Quality $\uparrow$} & \multicolumn{2}{c}{Geometric Consistency} & \multicolumn{2}{c}{Camera Accuracy} \\
    \cmidrule(lr){3-8} \cmidrule(lr){9-10} \cmidrule(lr){11-12}
    & & Subject & Background & Aesthetic & Imaging & Temporal & Motion & \multirow{2}{*}{Dyn-MEt3R $\uparrow$} & \multirow{2}{*}{MEt3R $\downarrow$} & \multirow{2}{*}{TransErr $\downarrow$} & \multirow{2}{*}{RotErr $\downarrow$} \\
    & & Consistency & Consistency & Quality & Quality & Flickering & Smoothness & & & & \\
    \midrule
    
    Redirector \citep{park2025redirector} & 1.0 & 0.9130 & 0.9143 & 0.5168 & 0.6755 & 0.9540 & 0.9857 & 0.8555 & 0.3206 & 0.0140 & 1.3689 \\
    Ours & 1.0 & \textbf{0.9178} & \textbf{0.9174} & \textbf{0.5191} & \textbf{0.6782} & \textbf{0.9552} & \textbf{0.9862} & \textbf{0.8638} & \textbf{0.3150} & \textbf{0.0123} & \textbf{1.2156} \\
    
    \midrule
    
    Redirector \citep{park2025redirector} & 1.5 & 0.8922 & 0.9004 & 0.5085 & 0.6713 & 0.9486 & 0.9838 & 0.8368 & 0.3354 & 0.0157 & \textbf{1.5476} \\
    Ours & 1.5 & \textbf{0.8983} & \textbf{0.9035} & \textbf{0.5111} & \textbf{0.6741} & \textbf{0.9495} & \textbf{0.9844} & \textbf{0.8458} & \textbf{0.3307} & \textbf{0.0129} & 1.5491 \\
    
    \midrule
    
    Redirector \citep{park2025redirector} & 2.0 & 0.8726 & 0.8881 & 0.4984 & 0.6624 & 0.9457 & 0.9820 & 0.8249 & 0.3496 & 0.0161 & 1.9246 \\
    Ours & 2.0 & \textbf{0.8783} & \textbf{0.8911} & \textbf{0.5017} & \textbf{0.6668} & \textbf{0.9469} & \textbf{0.9828} & \textbf{0.8338} & \textbf{0.3444} & \textbf{0.0153} & \textbf{1.8821} \\
    
    \bottomrule
  \end{tabular}
  }
\end{table}

\subsection{Evaluation protocol}

We follow the evaluation protocol of ReDirector \citep{park2025redirector}, using 50 videos from the DAVIS dataset. By applying 10 ReCamMaster \citep{bai2025recammaster} camera trajectories per video, we construct 500 test cases with lengths varying from tens to nearly a hundred frames. We restrict TrajectoryCrafter \citep{yu2025trajectorycrafter} and CogNVS \citep{chen2025reconstruct} to a maximum of 49 frames to prevent performance degradation; for the other methods, the evaluated frame length matches the dataset defaults.

For our metrics, we use ViPE \citep{huang2025vipe} to extract camera parameters to compute TransErr and RotErr. We also apply MEt3R \citep{asim2025met3r} for input video consistency, Dyn-MEt3R \citep{park2025steerx} for geometric consistency, and VBench \citep{huang2024vbench} for comprehensive aesthetic evaluation. Moreover, to evaluate complex trajectories, we compare our method with our base model (ReDirector \citep{park2025redirector}) under different camera speeds. Since large camera movements can produce consecutive featureless frames (e.g., sky, water or non-textured ground) that lead to ViPE \citep{huang2025vipe} estimation failures, we perform this speed-wise comparison on a reliable subset of 40 DAVIS \citep{pont20172017} videos to ensure fairness, applying 10 ReCamMaster \citep{bai2025recammaster} camera trajectories per video.

\begin{table}
  \caption{\textbf{Qualitative ablation results on DAVIS \citep{pont20172017} dataset.} Beyond improving geometric consistency and camera accuracy, the geometry reward also yields improvements in visual quality.}
  \label{tab:ablation_study}
  \centering
  \renewcommand{\arraystretch}{1.1} 
  \resizebox{\textwidth}{!}{
  \begin{tabular}{lcccccccccc}
    \toprule
    \multirow{3}{*}{Method} & \multicolumn{6}{c}{Visual Quality $\uparrow$} & \multicolumn{2}{c}{Geometric Consistency} & \multicolumn{2}{c}{Camera Accuracy} \\
    \cmidrule(lr){2-7} \cmidrule(lr){8-9} \cmidrule(lr){10-11}
    & Subject & Background & Aesthetic & Imaging & Temporal & Motion & \multirow{2}{*}{Dyn-MEt3R $\uparrow$} & \multirow{2}{*}{MEt3R $\downarrow$} & \multirow{2}{*}{TransErr $\downarrow$} & \multirow{2}{*}{RotErr $\downarrow$} \\
    & Consistency & Consistency & Quality & Quality & Flickering & Smoothness & & & & \\
    \midrule
    
    baseline (Redirector \citep{park2025redirector}) & 0.9098 & 0.9150 & 0.5141 & \underline{0.6821} & 0.9537 & 0.9856 & 0.8497 & 0.3130 & 0.0149 & \underline{1.4635} \\
    Video quality reward & \underline{0.9130} & \underline{0.9174} & \underline{0.5145} & 0.6818 & \textbf{0.9549} & \underline{0.9861} & \underline{0.8531} & \underline{0.3117} & \underline{0.0147} & 1.6082 \\
    Full reward & \textbf{0.9147} & \textbf{0.9181} & \textbf{0.5163} & \textbf{0.6828} & \underline{0.9548} & \textbf{0.9863} & \textbf{0.8550} & \textbf{0.3082} & \textbf{0.0140} & \textbf{1.3895} \\
    
    \bottomrule
  \end{tabular}
  }
\end{table}

\subsection{Main Results}

Table~\ref{tab:performance_comparison} demonstrates the effectiveness of our reinforcement learning framework. Compared to the baseline \citep{park2025redirector}, our RL-trained model exhibits notable improvements in Geometric Consistency, Camera Accuracy, and overall video quality. The gains in Geometric Consistency and Camera Accuracy are particularly substantial, validating that the incorporation of the geometry reward successfully guides the model to follow target trajectories with higher precision. Furthermore, Table~\ref{tab:camera_speed_comparison} presents the evaluation results across varying camera speeds. Our model consistently outperforms the baseline under more complex target trajectories, clearly illustrating the enhanced robustness across diverse trajectory conditions achieved through our RL training.

Qualitatively, Figure~\ref{fig:Qualitative results} visualizes our results on the DAVIS \citep{pont20172017} dataset. Our method demonstrates superior capabilities in maintaining geometric consistency between the foreground subject and the background. Notably, under large camera motions, baseline methods such as ReCamMaster \citep{bai2025recammaster} and ReDirector \citep{park2025redirector} frequently suffer from severe degradation, including subject disappearance and background blurring. In contrast, our approach significantly mitigates these collapse scenarios, robustly preserving both subject and background details. We attribute this enhanced stability to our RL training paradigm, which effectively exposes the model to a broader and more complex distribution of camera trajectories.

\subsection{Ablation Study}

To validate the effectiveness of our proposed mechanisms, we conduct an ablation study by retraining the model under different reward configurations. All models are trained for 140 steps on 16 A800 GPUs and evaluated against the baseline (ReDirector \citep{park2025redirector}) using the identical evaluation protocol and dataset as in Table~\ref{tab:performance_comparison}. Specifically, we compare the full reward formulation (incorporating both aesthetic and geometry rewards) against an ablated variant trained exclusively with the aesthetic reward. As shown in Table~\ref{tab:ablation_study}, relying solely on the aesthetic reward yields marginal overall performance improvements and even leads to a noticeable degradation in rotation accuracy. In contrast, integrating the geometry reward not only substantially enhances camera accuracy but also contributes to better visual quality and geometric consistency.

\section{Conclusion}

In this work, we introduce Geo-Align, a reinforcement learning framework designed for camera-controlled video retake. First, we design a reward mechanism using a metric 3D evaluator to explicitly optimize how accurately the generated video follows the target camera trajectory. Second and crucially, we propose a hybrid data sampling strategy combining real videos and scaled camera trajectories from synthetic data. This effectively mitigates the field's reliance on scarce time-synchronized multi-view video data, allowing the model to train on far more diverse scenes and complex trajectories. Empirical results validate that our approach consistently enhances camera control precision, consistency, and visual quality over existing baselines.





{
\small

\bibliographystyle{unsrtnat} 
\bibliography{reference.bib}
}


\appendix

\section{Appendix}

\subsection{Limitations}

Our reinforcement learning approach improves the accuracy of camera trajectory adherence in generated videos while simultaneously enhancing overall video quality. However, the model remains susceptible to failure when faced with excessively fast rotations, large translations, or large foreground objects close to the camera. Furthermore, inputs dominated by dynamic objects frequently lead to artifacts, such as the flickering or vanishing of those dynamic elements. Thirdly, the RL training process is highly time-consuming because it requires sampling the model to generate multiple complete videos for each batch, and the video generation process itself is inherently slow. Therefore, accelerating the RL training process remains a compelling direction for future exploration.

\subsection{Assets and Licenses}
\label{appendix:assets}

We summarize the assets used in our research, including their licenses and accessibility, in Table~\ref{tab:assets}. All assets are used in accordance with their respective terms.

\begin{table}[h]
    \caption{Summary of used assets (datasets, models, and code).}
    \label{tab:assets}
    \centering
    \renewcommand{\arraystretch}{1.1} 
    \resizebox{\textwidth}{!}{
        \begin{tabular}{lccc} 
            \toprule
            \textbf{Asset} & \textbf{Citation} & \textbf{License} & \textbf{Link/Source} \\ \midrule
            ReCamMaster & \citep{bai2025recammaster} & MIT License & \url{https://github.com/KlingAIResearch/ReCamMaster} \\
            ReDirector & \citep{park2025redirector} & Apache License 2.0 & \url{https://github.com/byeongjun-park/ReDirector} \\
            TrajectoryCrafter & \citep{yu2025trajectorycrafter} & Apache License 2.0 & \url{https://github.com/TrajectoryCrafter/TrajectoryCrafter} \\
            MixGRPO & \citep{li2025mixgrpo} & Non-Commercial License v1.1.1 & \url{https://github.com/Tencent-Hunyuan/MixGRPO} \\
            MapAnything & \citep{keetha2025mapanything} & Apache License 2.0 & \url{https://github.com/facebookresearch/map-anything} \\
            OmniWorld & \citep{zhou2025omniworld} & BY-NC-SA 4.0 & \url{https://github.com/yangzhou24/OmniWorld} \\
            CityWalk & \citep{li2025sekai} & No commercial license & \url{https://github.com/Lixsp11/sekai-codebase} \\
            ViPE & \citep{huang2025vipe} & Apache License 2.0 & \url{https://github.com/nv-tlabs/vipe} \\
            MEt3R & \citep{asim2025met3r} & MIT License & \url{https://github.com/mohammadasim98/met3r} \\
            Dyn-MEt3R & \citep{park2025steerx} & CC BY 4.0 & \url{https://github.com/byeongjun-park/SteerX} \\
            VBench & \citep{huang2024vbench} & Apache License 2.0 & \url{https://github.com/Vchitect/VBench} \\
            \bottomrule
        \end{tabular}
    } 
\end{table}





\begin{figure}
  \centering
  \includegraphics[width=0.99\linewidth]{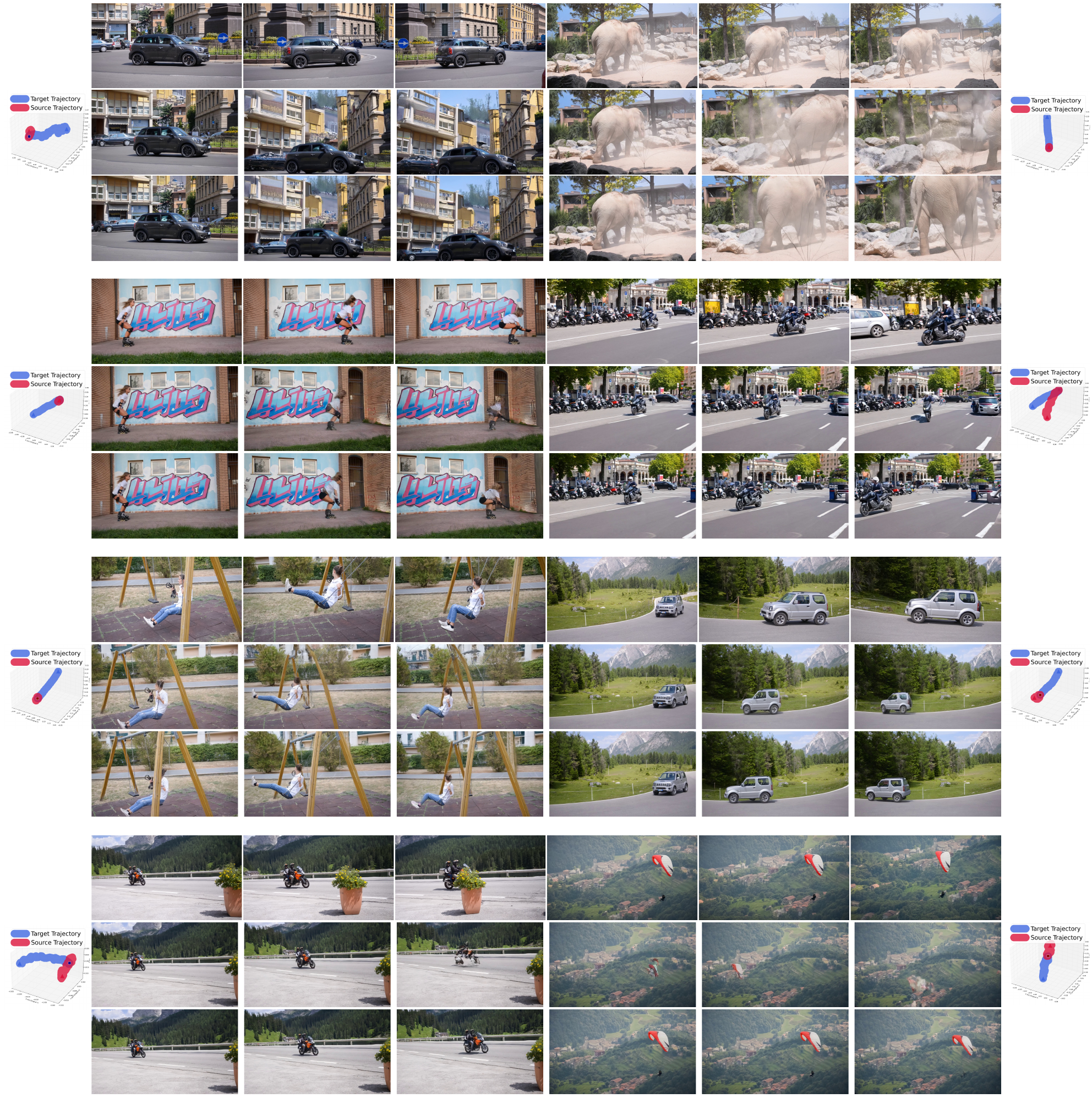}
  \caption{\textbf{More qualitative comparison on DAVIS \citep{pont20172017} dataset.} For each example, the top row illustrates the input video, while the second and third rows present the results of ReDirector \citep{park2025redirector} and our model, respectively.}
  \label{fig: more qualitative results}
\end{figure}

\begin{figure}
  \centering
  \includegraphics[width=0.99\linewidth]{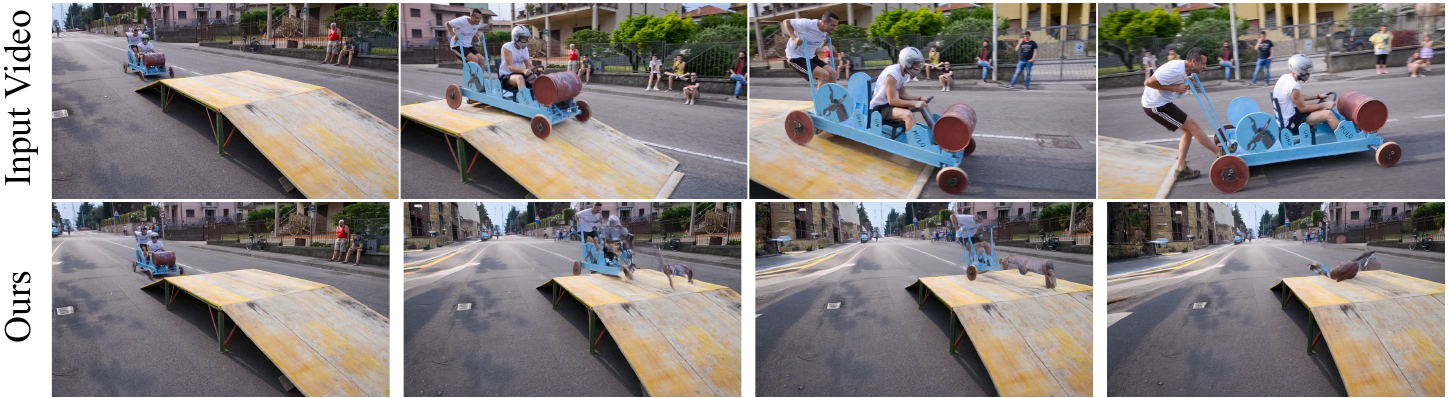}
  \caption{\textbf{Failure Case.} the model remains susceptible to failure when faced with excessively fast rotations, large translations, or large foreground objects close to the camera.}
  \label{fig: failure case}
\end{figure}



\end{document}